# A Bioinspired Bidirectional Stiffening Soft Actuator for Multimodal, Compliant, and Robust Grasping

Jianfeng Lin[1], Ruikang Xiao[1], Miao Li[2], Xiaohui Xiao[1], Zhao Guo[1*]

**Abstract—** The stiffness modulation mechanism for soft robotics has gained considerable attention to improve deformability, controllability, and stability. However, for the existing stiffness soft actuator, high lateral stiffness and a wide range of bending stiffness are hard to be provided at the same time. This paper presents a bioinspired bidirectional stiffening soft actuator (BISA) combining the air-tendon hybrid actuation (ATA) and a bone-like structure (BLS). The ATA is the main actuation of the BISA, and the bending stiffness can be modulated with a maximum stiffness of about 0.7 N/mm and a maximum magnification of 3 times when the bending angle is 45 deg. Inspired by the morphological structure of the phalanx, the lateral stiffness can be modulated by changing the pulling force of the BLS. The lateral stiffness can be modulated by changing the pulling force to it. The actuator with BLSs can improve the lateral stiffness about 3.9 times compared to the one without BLSs. The maximum lateral stiffness can reach 0.46 N/mm. And the lateral stiffness can be modulated decoupling about 1.3 times (e.g., from 0.35 N/mm to 0.46 when the bending angle is 45 deg). The test results show the influence of the rigid structures on bending is small with about 1.5 mm maximum position errors of the distal point of actuator bending in different pulling forces. The advantages brought by the proposed method enable a soft four-finger gripper to operate in three modes: normal grasping, inverse grasping, and horizontal lifting. The performance of this gripper is further characterized and versatile grasping on various objects is conducted, proving the robust performance and potential application of the proposed design method.

# 1 Introduction

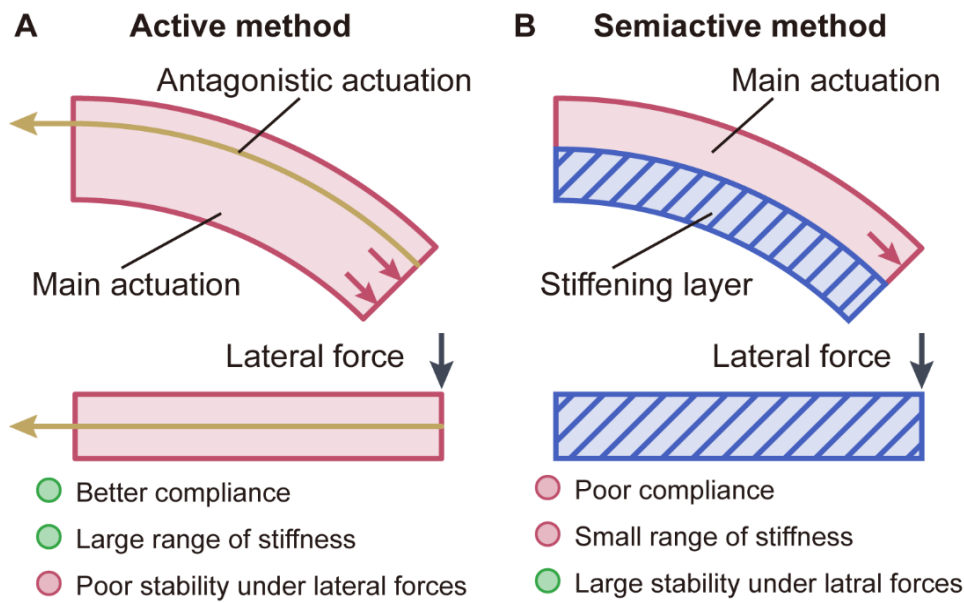

**FIG. 1.** The simplification schematic of two different stiffness modulation methods. (A) The active method is based on antagonistic manner. It has better compliance but the stability under lateral forces is poor. (B) The semiactive method controls the stiffness by changing the mechanical property of the stiffening layer. It can provide large stability, but the compliance and the range of stiffness modulation are reduced.

Soft robotics has gained increasing attention. Many real-world applications have been proposed due to its irreplaceable properties[1–18]: safety for interaction[19,20], adaptation to the environment[21,22], simplification to low-level control[23–25], and inspiration for novel bionic design[21,26–31]. However, the uncontrollable deformation and long-time response limit the applications due to the low stiffness[19,22,27].

Stiffening of soft tissues in nature enlightens the idea to overcome these difficulties: higher output force by selected stiffening trunk of elephants and higher locomotion efficiency of fish and legged animals via variable tail[32] and joint stiffness[24,25,33]. Stiffness modulation has also shown its engineering advantages in therapy exercises[17], wearable haptics[34], grasping and manipulation[1–5,7,9,22,35].

Among the existing approaches to modulating stiffness, there are two categories[27]: the active method based on an antagonistic manner (Figure 1A) and the semiactive method by changing the elastic properties (Figure 1B). In former approaches, two actuations are arranged antagonistically to stiffen the actuator[4,36–39]. For example, Li et al.[38] modulated stiffness through antagonism of different types of SMA. Yang et al.[40] designed a variable stiffness actuator based on pneumatic actuation and supercoiled polymer artificial muscles. Compared to the long response time of these two methods[2,41], Yang et al.[3] utilized a gas-ribbon-hybrid antagonistic arrangement to provide a fast-response variable stiffness solution.

For above mentioned active methods, a common shortcoming exists: poor stability under lateral forces. The antagonistic actuation (yellow arrow in Figure 1A) can only withstand tension but not bending. Therefore, the lateral stiffness depending on the low Young's modulus of the main body is low. Further applications are limited. For instance, compared to the rigid gripper, the soft gripper is unable to lift an object horizontally by utilizing lateral stiffness, which has been proven to be a more robust grasp strategy[42].

The semiactive method is to stiffen by changing the intrinsic mechanical properties of the stiffening layer (Figure 1B), including granular jamming[1,2,7,34,43,44], magnetorheological and electrorheological materials[45,46], and low melting point materials[47]. Several semiactive actuators demonstrate good performance in solving the low-lateral robust.

For example, Yan et al.[5] proposed a human-inspired finger that could tune the stiffness by heating the embedded conductive thermoplastic starch polymers (CTPSs) and its Yoshimura origami ligaments can also improve lateral stability. A soft-rigid hybrid gripper proposed by Li at el.[6] modulated the stiffness by heating the polylactic acid layer. Wei et al.[7] designed the variable stiffness gripper based on particle jamming. To improve the stability of the jamming mechanism, Jiang et al.[2] designed a chain-like jamming mechanism and its stiffness can be regulated by changing the pulling force.

For the above approaches, the lateral stiffness can be enhanced by an anisotropic stiffness structure or a thick rigid stiffening layer. However, the compliance of soft actuators and the range of tuneable bending stiffness are limited. For example, in the practical experiments, the actuator embedded with a chain-like structure[2] shows little range of stiffness modulation.

To briefly summarize, to obtain compliance, a large range of bending stiffness modulation while improving lateral stiffness is still a research gap in soft actuator design. Some similar works have also noticed this point: Bern et al.[48] proposed a skeleton-embedded soft finger, demonstrating stability and strength. However, the lateral stability of the finger and the ability to adjust the stiffness were ignored. Zhu et al.[49] utilized a soft-rigid hybrid design to implement a wide-range grasping force. By contrast, the rigid skeleton limits its omidirection's compliance.

To provide a compliant, large bending stiffness modulation range while having enhanced lateral stiffness to improve stability, we proposed a bioinspired bidirectional stiffening actuator (BISA) consisting of air-tendon hybrid actuation (ATA) and bone-like structure (BLS). ATA plays role in bending and stiffening in the bending direction. The BLS provides high lateral stiffness and maintains bending compliance at the same time. Meanwhile, the lateral stiffness can be changed by changing the pulling force to BLS.

Herein, the bioinspired design concept for BISA is detailed. We propose a dependable soft-rigid hybrid design and the corresponding fabrication method. Then the bending and lateral stiffness are analyzed to design the structure. To verify the analysis and the advantages, the characterizations of BISA are listed from three aspects: the coupling effect between two mechanisms, the bending stiffness, and the lateral stiffness. Finally, a multimodal gripper to perform versatile grasping was developed. Quantitative results and grasping tests both prove the advantages.

# 2 Design and Analysis

## 2.1 Design Concept

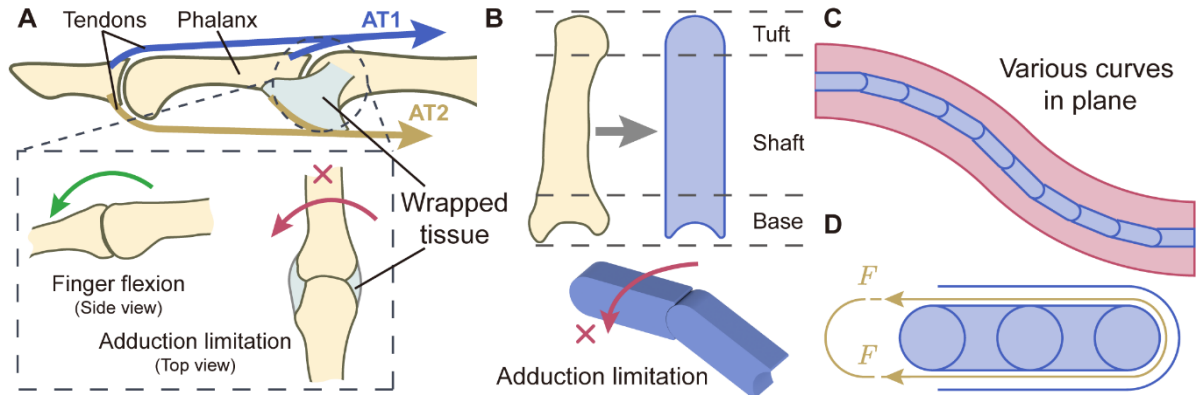

**FIG. 2.** Illustration of the design concept of the proposed mechanisms. (A) The Index finger anatomy shows the antagonistic manner, finger flexion, and adduction limitation. The adduction between two adjoint phalanges is limited by the wrapped tissue and the structure of the phalanx. (B) The bone-like structure is inspired by the phalanx. The tuft and base of the phalanx are simplified to a column. The adduction can be limited. (C) The bone-like structure can be adapted to various deformations in plane. (D) Schematic diagram of how BLS adjusts its stiffness and its connection method.

The anatomical features of the finger inspire the design in two aspects: the antagonistic tendons to adjust joint stiffness (AT1 and AT2, as Figure 2A illustrates), and the adduction limitation to provide high lateral stiffness. The bending stiffness of BISA can be modulated through the antagonistic manner of tendon and input air to improve grasping performance.

The bone-like structure (BLS) draws inspiration from the second feature to provide enhanced lateral stiffness while keeping compliance. As Figures 2A and B display, the engagement of the tuft and base of phalanges allows the finger to bend freely and prevents the finger from lateral deformation. On BLS, the tuft and base of a phalanx are considered a column with a constant radius. Rather than connective tissue that wraps around joint[48] or mechanical articulation[50], a tendon around each segment connects each joint as Figure 2D shows. Because the connective tissue provides less effective lateral stiffness due to the gap between joints and the mechanical articulation limits the structure compactness, compliance, and the ability to adjust the lateral stiffness[14].

The structure with multiple connected joints can adapt to complex deformation(as shown in Figure 2C). The lateral stiffness of BLS can be modulated by adjusting the tension to BLS, whose principle is similar to [2].

## 2.2 Structure Design

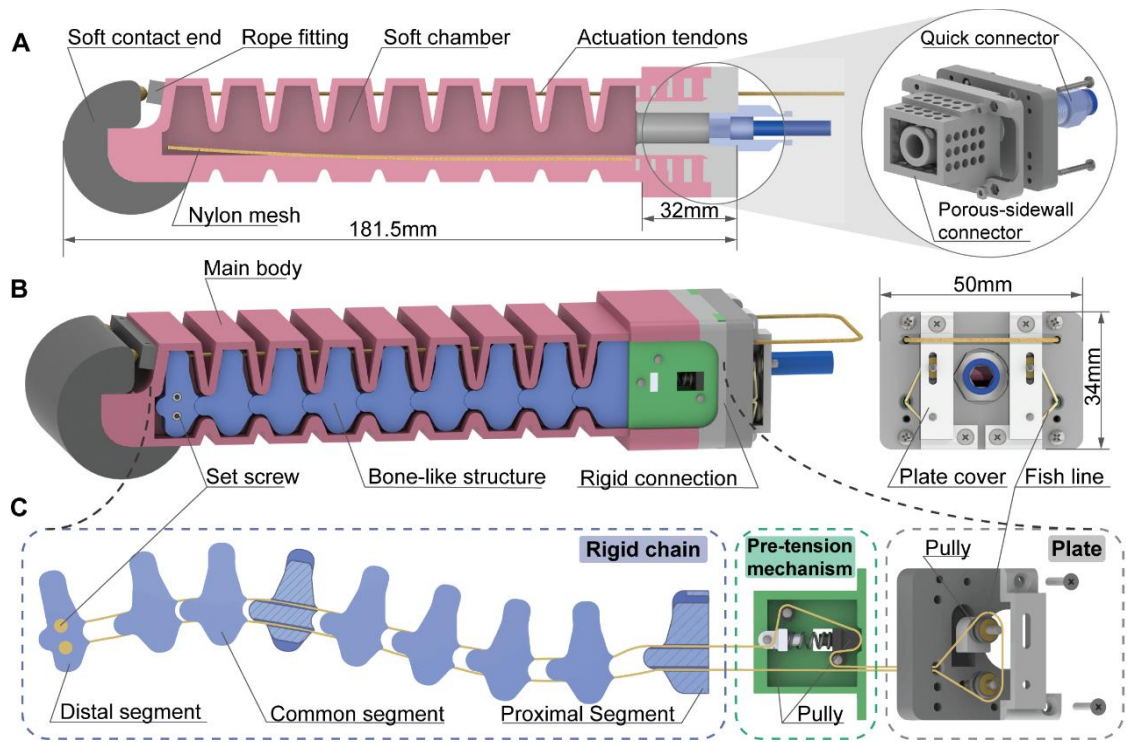

**FIG. 3.** The design details of the proposed actuator (BISA). (A) The sectional view of the main body of BISA and the details of the soft-rigid connection. (B) The overall structure diagram of BISA and the back view of BISA. The BLS is placed on each side of the main body, connecting to the rigid plate. (C) The details of the bone-like structure include three parts, the rigid chain, the pre-tension mechanism, and the plate.

The main structure of the BISA, as Figure 3B shows, is composed of three parts, the main body, bone-like structures (BLSs), and the rigid connection. Two BLSs are attached on each side of the main body, connecting to the connection part.

As Figure 3A displays, the main body includes three components: a soft contact end, the soft chamber, and the actuation tendons. A nylon mesh is inserted to limit the elongation of the main body (Smooth-On 0020 silicone with a 20-shore hardness). The soft contact end

(Smooth-On 0030 silicone with a 30-shore hardness), which is cylindrical is glued in the front of the soft chamber to increase the fingertip force[8]. The actuation tendons (Kevlar) cross through the holes which are in the same plane as the symmetrical plane of BLS and tie to the rope fittings at the end of BISA. The soft chamber is cast into the porous-sidewall connector. The silicone fills the gap and holes of the connector, enhancing the airtightness of the soft-rigid connection. The quick connector is connected to the connector. The details of fabrication can be found in Supplementary Note S1 and Supplementary Figure S1.

There are three major parts in each BLS as shown in Figure 3C: rigid chain, pre-tension mechanism, and plate. They are connected through a fishline going across them. The fishline is fixed by two set screws in the distal segment and its path is shown in Figure 3C in yellow. The bulge and concavity of the segments are semicylindrical, and the holes are perforated along the tangent of the cylinder. The segments are designed to match the size of the actuator to make the segment fit the main body. All the components of the pre-tension mechanism are 3D printed in nylon to bear a higher load. Two pulleys (nylon) on each side of BLS provide a path to guild the fishline. One of the pulleys is connected to the slider (PLA) placed in the chute of the backward plate (PLA).

There are two ways to modulate the lateral stiffness as Figure S2 shows: adding specific sizes of PLA blocks to change the position of the slider (Figure S2A) and directly controlling the pulling forces to BLS (Figure S2B).

### 2.3 Stiffness Analysis

The bending stiffness and lateral stiffness are analyzed separately to understand the variation of stiffness and provide theoretical guidance for the design.

2.3.1 Bending stiffness

The bending stiffness is modulated by the antagonism between air and tendons. When the tendon is slack, the actuator bends by the input air, enabling BISA for high compliance and adaptability. When the bending angle is controlled by input air and tightened tendon, the BISA can adjust bending stiffness in the position by incremental input air pressure.

Referring to study[51] and Supplementary Note S2, the bending stiffness modulation depends on the magnitude of the pressure and the number of cavities squeezing each other. The arrangement of the tendon has the advantage to increase the stiffness when the external load is along with the bending. Therefore, the number of chambers is 9 to improve modulation performance.

2.3.2 Lateral stiffness

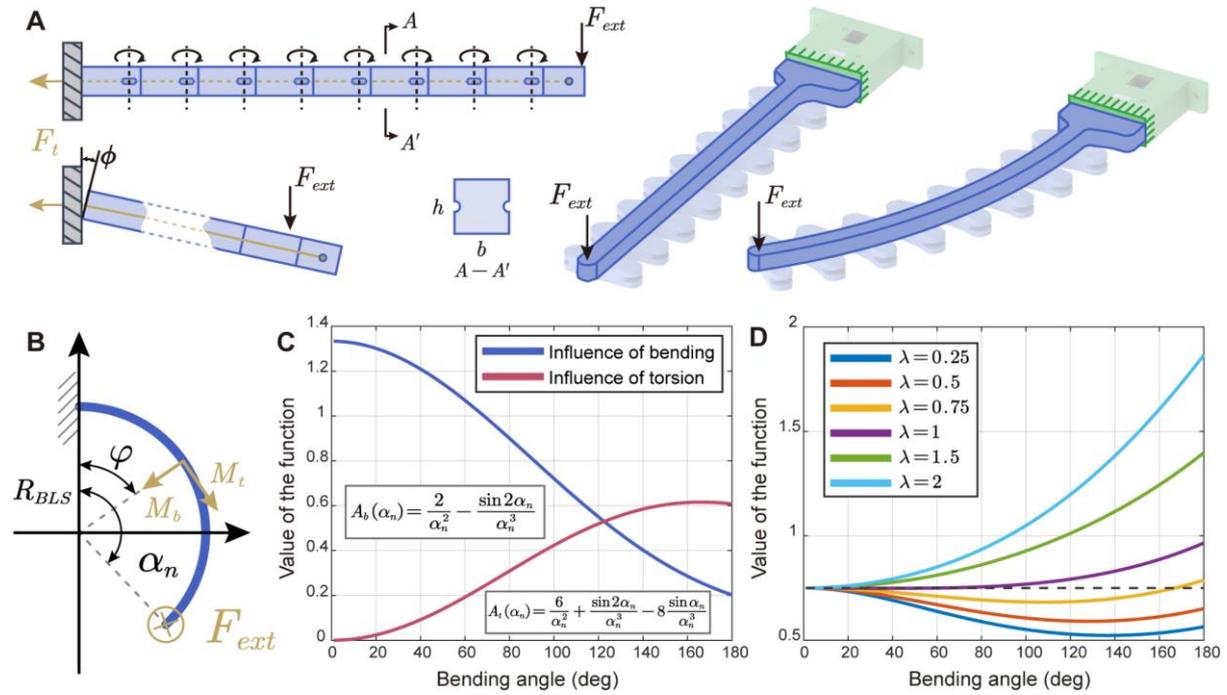

**FIG. 4.** Schematic diagrams and results about the analysis of factors influencing lateral stiffness. (A) The analysis of the stiffness of the bone-like structure. (B) Simplified model of the curved cantilever beam. (C) The influence of bending and torsion due to the external forces on deformation as the bending angle increases from 0deg to 180deg. (D) The trend of stiffness with angle for different aspect ratios.

The lateral stiffness can be changed by changing the polling force to BLS. Higher pulling force provides higher lateral stiffness. Referring to [2], as shown in Figure 5A, the BLS works normally when

$$F_t h \geqslant F_{ext} L \tag{1}$$

where $F_t$ is the pulling force provided by rope and $F_{ext}$ denotes the external force applied to the end of BLS, $h$ represents the height of BLS, and $L$ is the length of the rigid chain.

When the BLS works normally and bends at a certain angle, the BLS is simplified into a curved beam with a constant radius $R_{BLS}$ as Figures 5A, and B illustrate. The axial length of BLS is invariable, and the radius satisfies

$$R_{BLS}\alpha = C \tag{2}$$

where $\alpha$ is the bending angle of BLS, $C$ is a constant. According to Supplementary Note S3, the lateral stiffness of an equivalent beam can be derived from Castigliano's second theorem as

$$k = \frac{4EI}{C^3} F(\alpha) \tag{3}$$

where $E$ is Young's modulus, and $I$ refers to the area moment of inertia. $F(\alpha)$ is the evaluation function whose value varies with the bending angle:

$$F(\alpha_n) = 1 \bigg/ \bigg[ A_{bending}(\alpha_n) + \frac{2(1+\nu) A_{torsion}(\alpha_n)}{(1+\lambda^2)} \bigg] \tag{4}$$

where $\nu$ is the Poisson's ratio of the BLS's material and the $\lambda$ is the aspect ratio of the beam. The aspect ratio is $\lambda = h/b$, where $b$ is the width of the equivalent beam. The $A_{bending}(\alpha)$ and $A_{torsion}(\alpha)$ represent the influence of bending and torsion.

Figure 4C depicts the trend of the functions $A_{bending}(\alpha_n)$ and $A_{torsion}(\alpha_n)$, indicating that with the bending angle varying from 0° to 180°, the influence of torsion increases. The simplification of BLS into a continuous cantilever beam ignores the rope connection which has a low resident to torsion. Predictably, as the influence of torsion increases, the model will become more inaccurate.

The Poisson's ratio (around 0.35 [52]) and the aspect ratio varies from 0.25 to 2 are taken into the evaluation function, as shown in Figure 4D. When the aspect ratio is greater than 1, the lateral stiffness of BLS increase with the bending angle. Meanwhile, according to (4), the stiffness depends on the moment of inertia $I = \lambda b^4/12$. To make the performance dependable, the lateral stiffness of BLS should not decrease with bending, which means larger $\lambda$ and $b$ are needed. However, the effective region of BLS corresponds to the bending center of the soft chamber, limiting the size of BLS. Here, we choose $\lambda = 1$ to meet the basic requirements.

## 3 Experiments and Results

### 3.1 Bending Angle

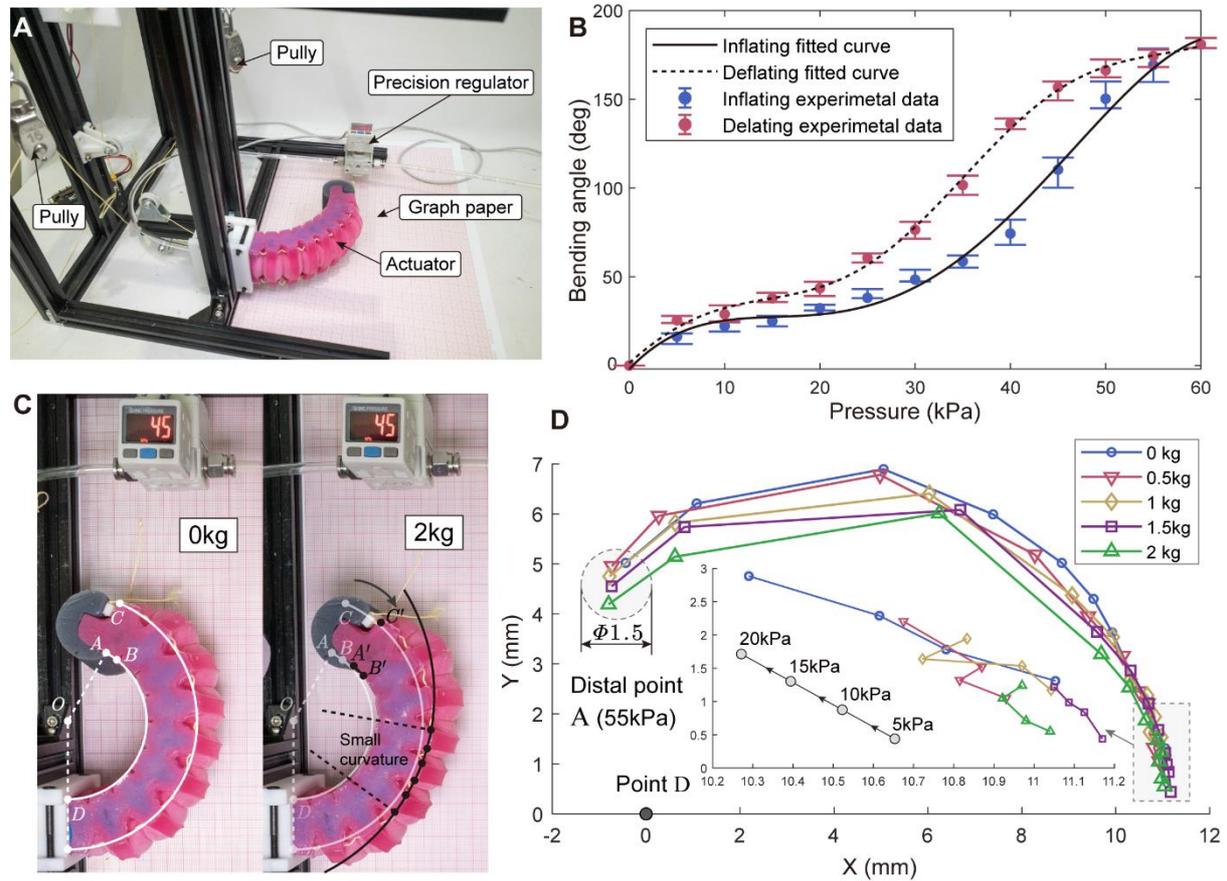

**FIG. 5.** Characterization of the influence of BLSs on bending. (A) The experiment platform. (B) The testing results of the bending angle vary with gas pressure. (C) Intuitive comparison of bending when the pulling force to BLSs is 0kg and 2kg.

The BLS should not influence the compliance of bending. To test the deformation and the coupling influence brought by the rigid structures, the experimental platform as Figure 5A shows was set. The actuator is connected to the precision regulator (IR1000-01, SMC Ltd., scale 5kPa to 200kPa, accuracy ±3%). The lateral stiffness of BLS was controlled in the second way as Figure S2B. Each BLS's proximal end was connected to a pully to connect with weights. The graph papers were placed under the frames to record positions.

Figure 5C shows an intuitive comparison to demonstrate the influence on bending. The two groups were both pressurized to 45 kPa while with different pulling forces, 0 kg and 2 kg. When the pull weight increased to 2 kg, the deformation was composed of several different segmented curvatures unlike the constant curvature arc when the pulling weight was 0 kg. The middle segment has a smaller curvature as Figure 5C shows. The distal point of BISA lagged that when the pulling weight was 0 kg.

To further evaluate the deformation, two experiments were conducted. Firstly, the bending angle when the pulling weight is 0 kg was tested. Points $A$ and $B$ were marked on the graph paper to obtain the bending angle, as shown in Figure 5C. To ensure safety, the BISA was inflated from 0 kPa to 60 kPa in 5 kPa steps and then deflated back. The average data were fitted after 5 repeated experiments. As Figure 5B illustrates, the errors between the inflating and deflating show the hysteresis behavior of the hyperelastic material[3], which is similar to the soft actuators without the inserted rigid structure.

Second, the position of the distal point $C$ was chosen to evaluate the influence of BLS on bending. Five groups of pulling forces were chosen: 0 kg, 0.5 kg, 1kg, 1.5 kg, and 2 kg. And the input pressure increased from 0 to 55 kPa. A minimum circle was calculated to cover the five recorded points $C$ and the center of the circle was determined to be the average position. Figure 5D displays all the positions of the distal point. With the increase of the pulling weights, the deformations share similar trends and the maximum position error is about 1.5 mm, which can be ignored in many practical scenes. As Figure 5D shows, the first few points in the condition with load are lagging. This may be because the bending moment is less than the resistance moment of the segments' friction

## 3.2 Stiffness Tests

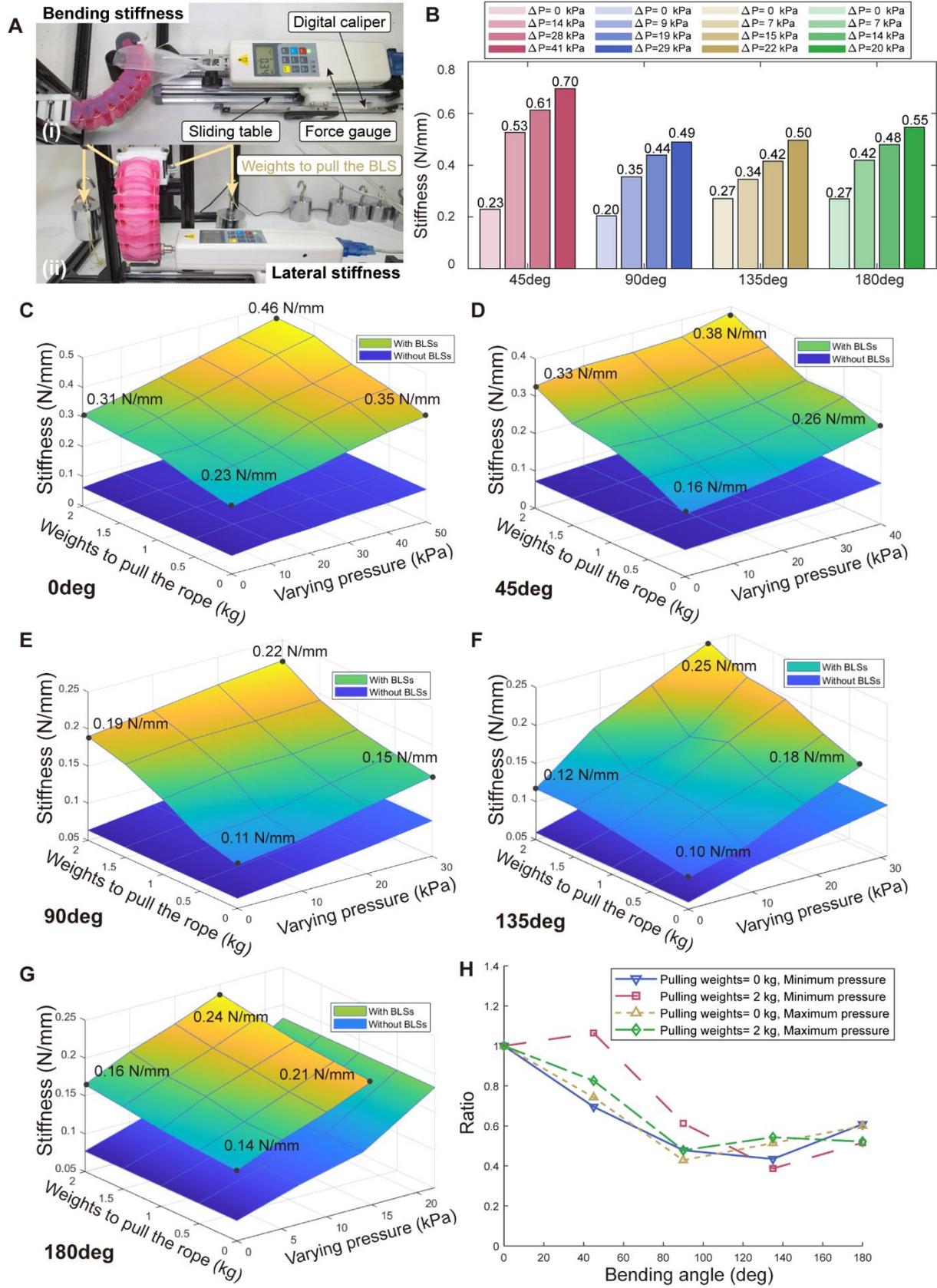

**FIG. 6.** Testing of the bending stiffness and lateral stiffness. (A) The experimental platform for (i) bending stiffness and (ii) lateral stiffness. (B) The summary of test results of bending stiffness. The testing results of lateral stiffness when bending angle is (C) 0deg, (D) 45deg, (E) 90deg, (F) 135deg, and (G) 180deg. (H) The change of stiffness ratio with the bending angle under four different input air and pulling force conditions.

To verify the bidirectional stiffening property, the experiments were designed as Figure 6A shows. The BISA was connected to a sliding table with a force gauge (HP-200, HANDPI Ltd., scale 0 - 200 N, accuracy ±0.5%) and digital caliper (HANDPI Ltd., scale 0 - 150 mm, accuracy ±3%). A brushless geared motor (M2006, DJI technology Ltd., rated torque 1N*m) controlled the tendon. The lateral stiffness of BLS was modulated through the second method (Figure S2B) by varying the weights to the ropes. For bending stiffness, as Figure 6A(i) illustrates, a carbon-fiber tube with a diameter of 26mm is connected through nylon mesh to the force gauge to contact the actuator. For lateral stiffness, the force gauge contacts the sidewall of the bending BISA directly. All experiments were repeated three times.

### 3.2.1 Bending Stiffness

To focus on testing the bending stiffness modulation, the pulling weight is 0 kg. The direction of the external force was specified as the extension direction of the actuator as Figure 6A(i) and S4A-D show. Four groups of bending angles were experimented: 45°, 90°, 135°, and 180°. The initial input pressure when the actuator bent exactly to a certain angle was recorded as $\Delta P = 0 kPa$. Four different incremental pressures were tested. When the bending angle was 90°, 135°, or 180°, the sliding table moved rightward 10mm slowly 1mm per step and 7mm rightward for 45° due to the unstable deformation when the slide moved larger than 8mm.

The average results of the four experiments are shown in Figure S4A-D. The relationship between the pulling force and the displacement is nearly linear. Figure 6B lists the average slope of each fitted curve. In summary, for different bending angles, 45°, 90°, 135°, and 180°, the bending stiffness is improved by about 3.0, 2.4, 1.8, and 2.0 times, respectively. The maximum stiffness can reach 0.7N/mm when the bending angle is 45° and the increased air pressure is 41kpa. These results verify the effectiveness of the tendon-air hybrid mechanism in bending stiffness modulation and indicate that the modulation range of stiffness is limited by the maximum allowable pressure.

The results when the bending angles are 90°, 135°, and 180° are similar. The layout of experiments contributes to this phenomenon. As Figures S4C and D show, in the 135° and 180° groups, the tube contact with BISA at the quarter-arc position to ensure stability. Therefore, not all the chambers were squeezed. The amount of contact moment $M_c$ (equation (1) on Supplementary Note S2) is smaller than when all the chambers can squeeze each other.

### 3.2.2 Lateral stiffness

The lateral stiffness of BISA was co-determined by the input pressure and the pull weights to BLSs, varying with the bending angle of BISA. Therefore, the experiments which were divided into five different bending angles (0°, 45°, 90°, 135°, and 180°) have two experimental variables: the pulling force to BLS (from 0kg to 2kg by 0.5kg each step) and input pressures (at least four steps). The lateral stiffness of BISA without BLSs was also tested. The same air pressure recording protocol as the last experiment was used here. The sliding table moved leftward at a 1 mm interval from 0 to 10mm for each case. The stiffness is calculated from the average slope.

Figure 6C-G list all the testing results of lateral stiffness. Compared with the cases without BLSs, the BLSs have efficiently improved the lateral stiffness with a maximum magnification of 3.9 times when the bending angle is 45°. As can be seen from Figure 6C-G, the lateral stiffness can be modulated decoupling by adjusting the weights of the BLSs and the maximum adjustable multiplier reaches 1.9 times. The ability to tune the lateral stiffness with the pulling weights has been verified.

The lateral stiffness of BLS was designed to increase with the bending angle. Here, the variations of four different conditions with bending angles are depicted in Figure 6H. The ratios of stiffness relative to the value when the bending angle is 0 are calculated. The graph shows that there has been a slight decrease followed by a rising, which does not match the design goal but is close to the trend of the analysis model when the aspect ratio $\lambda$ is smaller than 1. Several factors contribute to this. First, the model used for analysis is rough for ignoring the easy-torsion property of BLS. Second, the equivalent Poisson's ratio of the beam is not precise considering the elastic fishline between segments.

As Figure 6C-G and Figure S5 illustrate, the stiffness magnification is influenced by input pressure. Figure S5 shows the lateral stiffness magnification with the pressurization steps for different bending angles. When the incremental air is the largest, the regulation of lateral stiffness is often the least effective. These results are likely to be related to the axial elongation of BISA. With the increase of input pressure, there is still a minor axial elongation even though a nylon mesh is inserted. The elongation leads to increased space between segments of BLS, resulting in low lateral stiffness.

### 3.3 Application on grasping

To demonstrate the advantages in applications, a four-finger gripper was designed and tested. The design details can be found in Figure S6 and Supplementary Note S4. The lateral stiffness of each BISA is controlled by the first method as Figure S2A. Each BLS is pre-tensioned to about 10N to improve the lateral stiffness passively.

### 3.3.1 Characterization of the Gripper

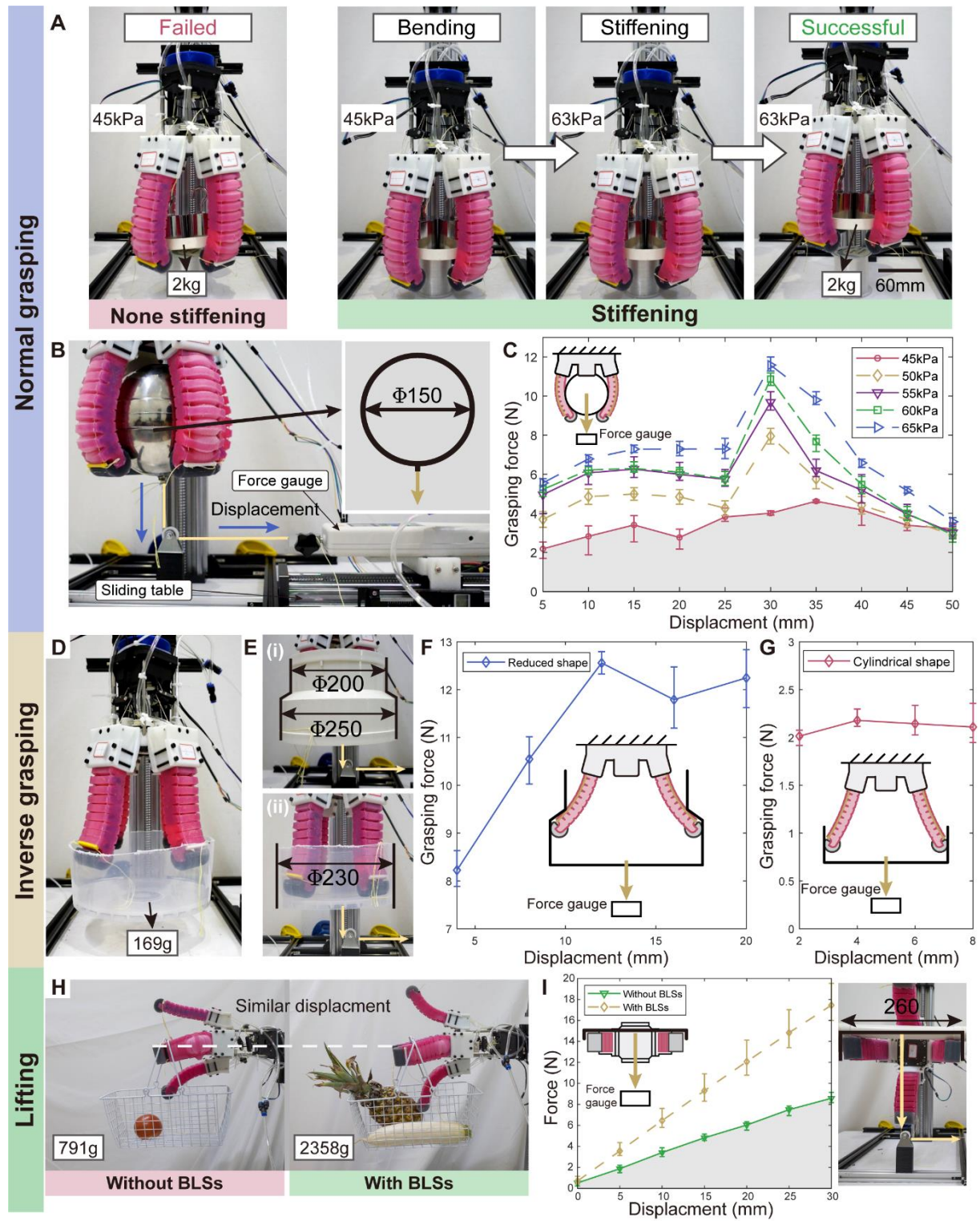

**FIG. 7.** Grasping performance in three modes. Normal grasping: (A) the intuitive comparison between none-stiffening and stiffening grasping, (B) the experimental set to test (C) the variation of grasping force with displacement under different increased input pressures. Inverse grasping: grasping

(D) half bucket with 169g. (E) Experiments to test the inverse grasping forces in two conditions: (i) cylindrical shape and (ii) reduced shape. The inverse grasping forces on (F) reduced shape and (G) cylindrical shape. Lifting: (H) comparison between the gripper with BLSs and without BLSs. (I) The lifting forces of the gripper with BLSs and without BLSs.

The soft gripper was designed to be operated in three modes: normal grasping, inverse grasping, and horizontal lifting. For normal grasping, the tuneable bending stiffness plays a role in performing higher stability and grasping force. Figure 7A and Supplementary Video S1 show a comparison between stiffening grasping and none stiffening grasping. The gripper failed to grasp the plate with 2 kg weights without stiffening. On contrary, the stiffening finger with the same posture grasps the plate successfully.

The platform as depicted in Figure 7B was designed to quantify the grasping force. When the input pressure was 45 kPa (none stiffening grasping state), the tendons were relaxed. In the stiffening state, the tendons were tightened when the input pressure varied from 50 kPa to 65 kPa. The grasping force of the stiffening state has been improved significantly. The maximum gripping force is increased by nearly three times, indicating the advantages of bending stiffness modulation.

The inverse grasping utilizes the reverse-driven tendon, ignoring the negative pressure to better clarify the role of the tendons. Figure 7D and Supplementary Video S3 display the successful grasping of a half bucket (169g). As Figure 7E shows, the inverse grasping forces on two different shapes were also tested: a reduced shape with diameters of 200 and 250mm (Figure 7E-i) and a cylindrical shape with a 230mm diameter (Figure 7E-ii). The testing results (Figure 7F-G) prove that the tightened tendon helps the inverse grasping. The grasping force on a cylindrical shape is less than that on a reduced shape because the inverse grasping on a cylindrical shape depends on friction.

An Intuitive comparison between the gripper with BLSs and without BLSs is shown in Figure 7H to highlight its improvement in lifting strength. Two fingers shared similar displacement while the one with BLSs endured a larger load of about 3 times. Figure 7I for the quantitative experiment of lifting force also shows the advantages. The maximum force of the lifting with BLSs can reach about 17N and the stiffness of this is about 2 times compared to that without BLSs.

### 3.3.2 Versatile Grasping Demonstration

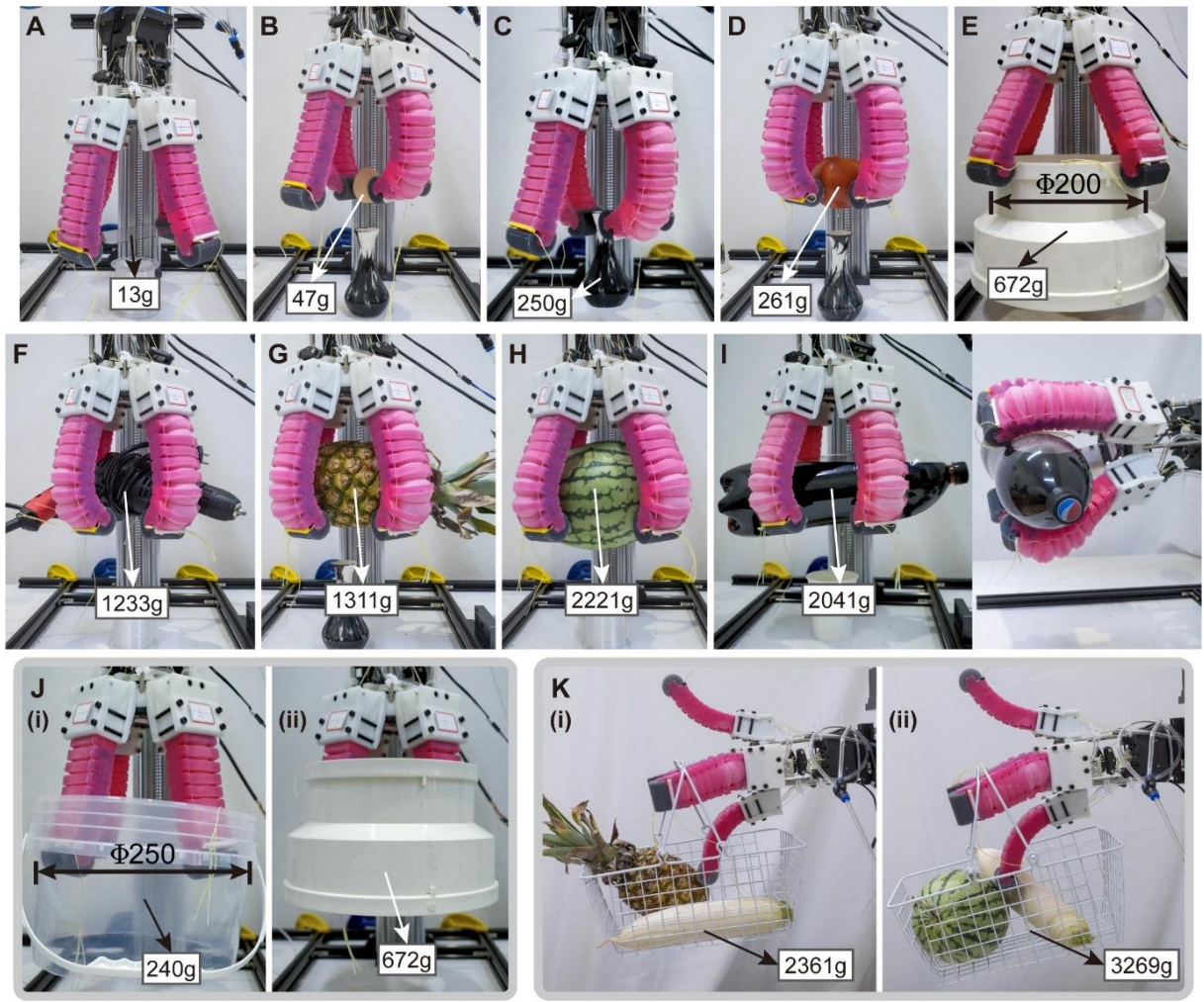

**FIG. 8.** Demonstration of versatile grasping. Pinching the targets including (A) a mobile phone film, (B) an egg, (C) a ceramic, (D) a tomato, and (E) a reduced pipe. Power grasping targets: (F) an electric drill, (G) a pineapple, (H) a watermelon, and (I) a drink with 2041g weights. (J) Inverse grasping (i) half bucket and the (ii) reduced pipe. (K) Lifting a basket with (i) 2361g weights and (ii) 3269g weights.

As shown in Figure 8, BISA enables versatile grasping tasks including multi-scale and multi-weight objects. All the targets were grasped successfully as Figure 8 displays and the dynamic process can be found in Supplementary Video 2-4.

Although with solid components, BISA shows its compliance. The gripper succeeds in pinching the fragile stuff like the mobile phone film (13g), egg (47g), ceramic (250g), and tomato (261) as Figure 8A-D illustrate. The gripper also shows its strength by grasping the electric drill (1233g, Figure 8F), pineapple (1211g, Figure 8G), and watermelon (2221g, Figure 8H). Due to the compliance and stiffness modulation, the gripper is capable of pinching a reducer water pipe with a large diameter (672g, and with a diameter of 200mm, Figure 8E). Even more, a bottle of drink weighing 2041g was successfully picked up when the gripper was set horizontally and vertically (Figure 8I). When the gripper was set horizontally, the enhanced lateral stiffness shared some of the stress making the grasp successful.

The gripper also conducted several inverse grasping and horizontal lifting tasks as shown in Figure 8J-K. A half bucket (240g, with a diameter of 250mm, Figure 8J-i) and a reducer pipe (672g, with diameters of 200mm and 250mm, Figure 8J-ii) were successfully grabbed up from the inside. As Supplementary Video S2 shows, the compliance allowed the gripper to be sent into the orifice and subsequently inverse driven to a larger diameter to lift the target. For the horizontal lifting, with the help of BLSs, the buskets carrying 2361g (Figure 8K-i) and 3269g (Figure 8K-ii) were carried vertically for a distance as shown in Supplementary Video S4 successfully.

In the above demonstrations, the versatility of grasping is achieved with the property of BISA. This initially validates the effectiveness of the proposed design.

## 4  Conclusion and Future Work

This paper presents a bioinspired bidirectional stiffening soft actuator (BISA) with a high and controllable lateral stiffness while a wide range of bending stiffness. The bending stiffness is successfully modulated by air-tendon-hybrid actuation (ATA). And the bone-like structure (BLS) contributes to the high and controllable lateral stiffness. The coupling influence between these two mechanisms is proved to be minor. A multimodal gripper with BISA is designed, including normal grasping, inverse grasping, and horizontal lifting. The characterizations and versatile grasping show the advantages of BISA in high grasping force, lifting force, and stability. Especially for the lifting mode, the gripper can lift a bucket with 3269g horizontally.

It can be concluded that our proposed design can provide a robust lateral performance meanwhile keeping the bending compliance and the bending stiffness modulation. The lateral stiffness can be tuned independently, enabling more potential uses in the future. Our future work will focus on the optimization of ATA and BLS, and further applications of the bidirectional stiffening property like in the rehabilitation glove and inchworm locomotion.

## Acknowledgments

Research supported by the National Natural Science Foundation of China (Grant No. 51605339), the Research Project of China Disabled Persons' Federation-on assistive technology (Grant No. 2021CDPFAT-27), the Key Research and Development Program of Hubei Province (Grant NO. 2020BAB133). We are grateful to the Wuhan University Student Engineering Training and Innovation Practice Center for the venue and equipments support.

## Author Disclosure Statement

No competing financial interests exist.

# Supplementary Figure

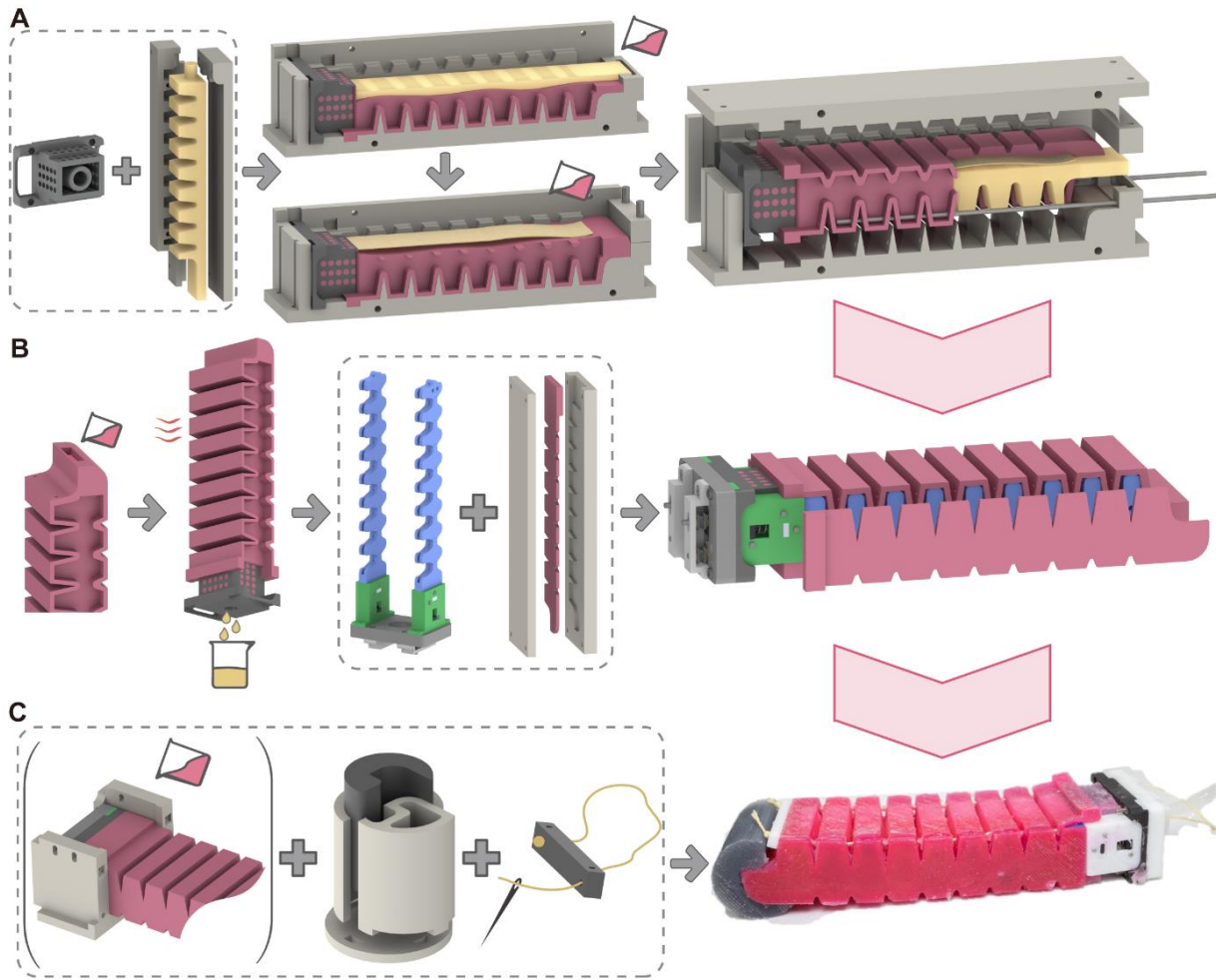

**FIG. S1.** The fabrication method of BISA. Firstly, (A) fabricate the soft chamber. Connect the wax core with the porous-sidewall connector and then moulded in the mold. Second, (B) eliminate the wax core by heating and fitting the BLSs to the main body. Third, (C) Assemble other components like rope fitting, actuation tendon, and the contact end.

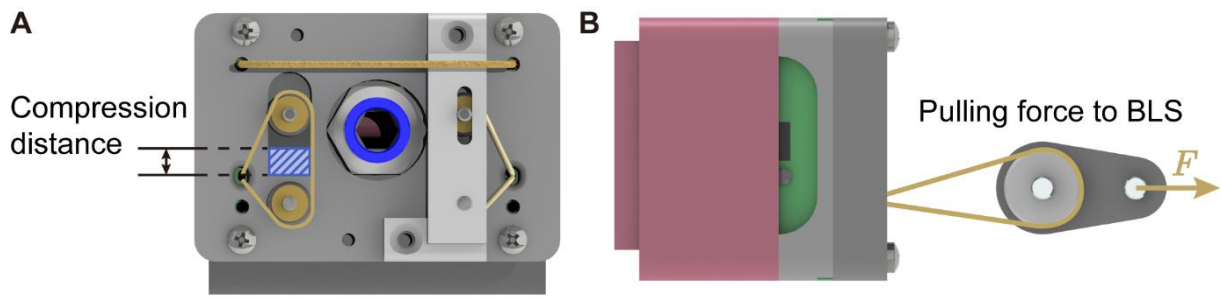

**FIG. S2.** The two stiffness modulation methods for BLS. (A) Add the PLA block (purple) to control the compression distance. (B) Connect the end of the rope to a pully and then control the pulling force to BLS directly.

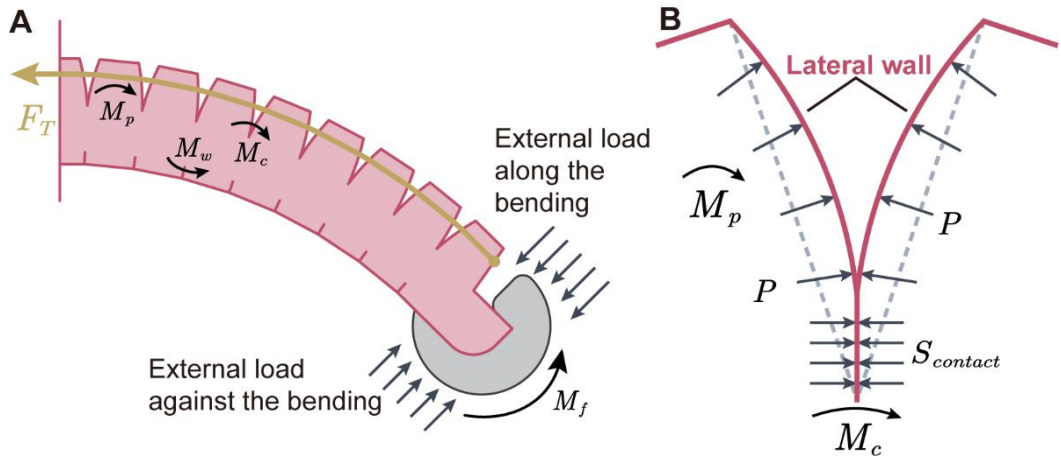

**FIG. S3.** (A) The moment and force of the MTSA. (B) The schematic diagram of the forces when adjacent cavities are in contact.

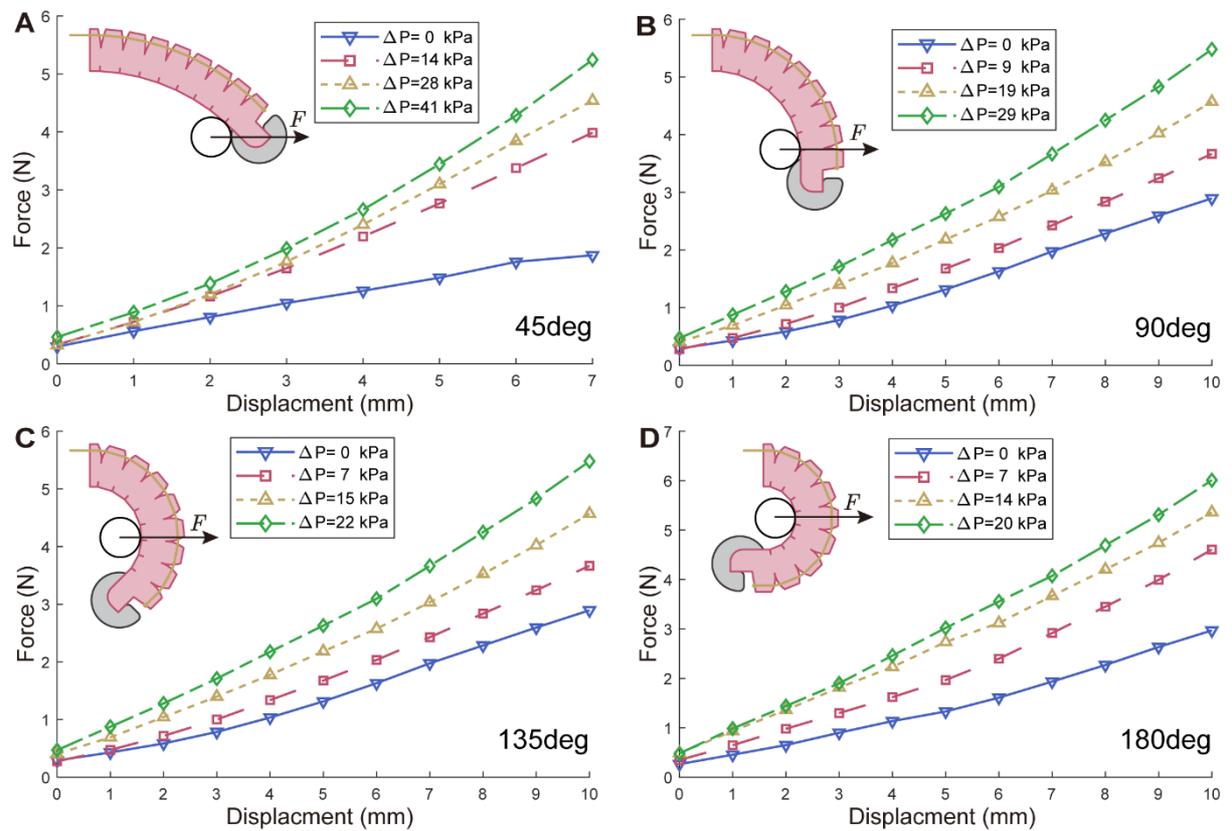

**FIG. S4.** The experiment results of bending stiffness for (A) 45deg, (B) 90deg, (C) 135deg, and (D) 180deg.

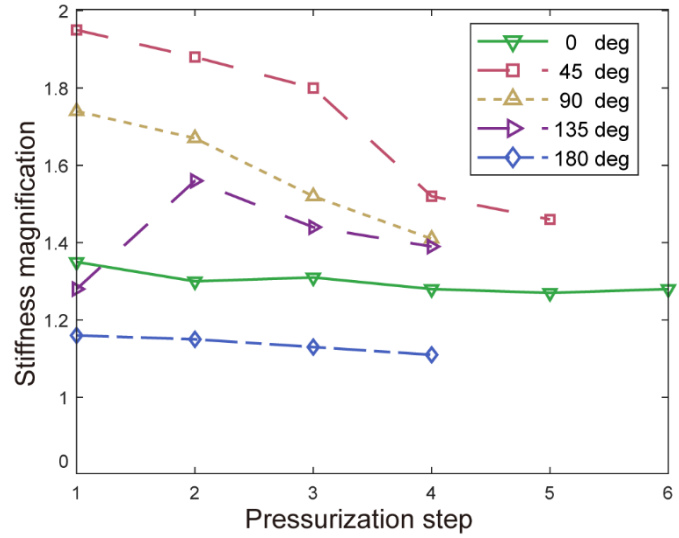

**FIG. S5.** The variation of stiffness magnification with increasing air pressure at the same angle

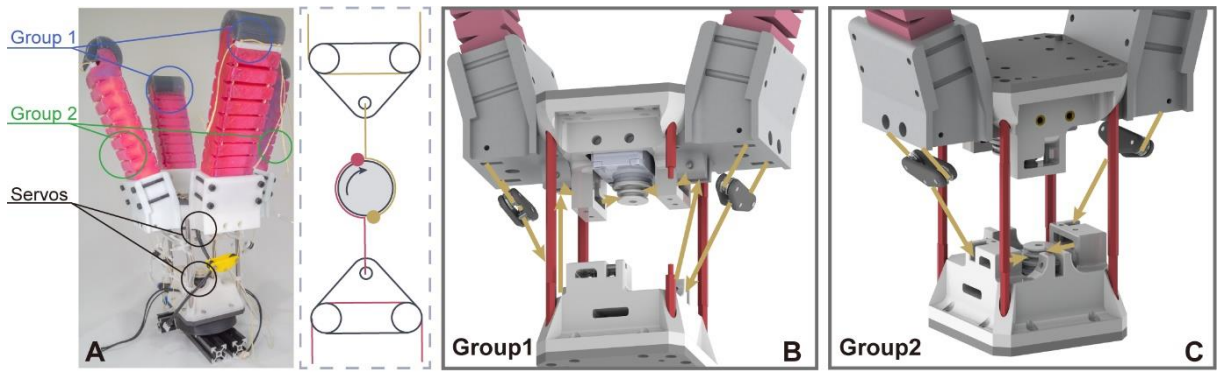

**FIG. S6.** The design of the proposed four-finger gripper. (A) Grouping of the four drives and actuation principle. (B) Alignment of group 1. (C) Alignment of group 2.

# Supplementary Note

## 1 Supplementary Note S1

To fabricate soft actuators with complex structure of inner chambers, the step-by-step casting methods were commonly used. However, the leakage may happen at the boundary of connection between layers[3], limiting the endurable input pressure. To avert this drawback, in this work, the BISA is fabricated based on combining the porous-sidewall connector and the lost-wax method. The lost-wax method ensures the one-piece modeling and silicon cured in the porous structure enhances the airtightness between the rigid component and soft material. The fabrication process, which is shown is Figure S1, includes three steps: fabrication of soft chamber, fitting the BLSs, and assembling other components. All the molds for casting were 3D printed with Nylon.

Step 1, fabrication of soft chamber. As Figure S1A displays, initially, to fabricate the wax core, the melted wax with a low melting point (about 70°C) was poured into the molds and the core would be demolded after the wax solidified. Then the wax core was fitted to the porous-sidewall connector through the cylindrical structure on the wax core. The assemble of the connector and core was placed in the mold after fluid silicone, which was a mixture of A and B parts at 1:1, have been poured into mold. Continuously pour the silicone and then add the nylon mesh on the top before closing the molds. When the silicone filled the molds, all parts of the molds were assembled through nuts and two metal sticks were inserted through holes on the bottom of molds. Finally, after curing in the dry oven at 65°C for 45min, the soft chamber with a wax core embedded was obtained. It is noteworthy the wax core was set on the mold as a cantilever beam with a simply supported to avoid the uneven wall thickness of soft chamber when the wax was placed as a cantilever beam.

Step 2, fitting the BLSs. To begin with, as Figure S1B shows, the stretching part of the wax core needed to be broken and then the silicone was poured into the space to get a complete actuator. Thenceforth, the soft chamber with wax core in it was heated in the water bath to 85°C for 25 mins. The molten wax flew down form the hole of connector. And then two pre-prepared BLSs connected on the plate were assembled with soft chamber by being fixed on the connector through nuts and screws. Subsequently, two cover layers were cast and bonded with the soft chamber by uncured silicone.

Step 3, assembling other components. In this step, the sealing layer for BISA was fabricated by placing the actuator in the mold and pouring the silicone into it as presented in Figure S1C. As for casting the contact, similar method was used. The contact end then was glued on the front of BISA by silicone rubber adhesive (Smooth-on, Sil-Poxy, USA). The actuation tendon was guide to crosse through the reserved holes by a needle and was fixed in the rope fitting. After connecting the quick connector to the porous-sidewall connector, the BISA was fabricated. It noted that the procedure to cast the sealing layer is not necessary, since the soft-rigid connection in porous structure has played the main role in reinforcing the airtightness, which we have discovered during our tests.

## 2 Supplementary Note S2

For bending, the stiffening of BISA is implemented through the antagonistic actuation of tendon driven and gas driven. To investigate the factors that influence the bending stiffness of BISA, two directions of the external load are discussed, respectively. Here we analyse the factors that stiffen the actuator from the point of view of the forces that block the deformation.

When the external load against the bending is applied to BISA, firstly, the tightened tendon will transfer to the relaxed state within a small displacement. The critical moment of the tendon in tightened state and relaxed state is noted as $M_{CR}$. Then, if the external moment ($M_f$) exceeds $M_{CR}$, the BISA will have a large deformation along the external force. Regarding the deformation process as a quasistatic one, according to Liu et al.[51], the equilibrium of moments for torque created by the external force ($M_f$), as Figure S3A shows, satisfies the following equation:

$$2(n-1)M_p + (n-1)M_c = 2(n-1)M_w + M_f \qquad (1)$$

where $M_p$ represents the moment due to the air pressure acting on the air chambers, $M_c$ is the contact moment of two adjoining chambers, and $M_w$ is the elastic restoring moment of inflated chambers, as Figure S3B illustrates. And the $M_p$, and $M_c$ can be calculated by the following equations:

$$\begin{cases} M_p = PSL = 4Pab^2 \\ M_c = PS_{contact} \end{cases} \qquad (2)$$

where the $S_{contact}$ is the contact area of two adjoined lateral walls. The $M_w$ is determined by the degree of stretching of the hyperelasticity material, which has little change in stiffening when the banding angle remains unchanged. Therefore, $M_w$ is treated as constant during stiffness modulation. Then, according to equation (1) and (2), the ability for BISA to withstand the external load, can be treated as stiffness is dependent on the input pressure and the contact area.

When the external load is along the bending, the air pressure dose not resist the external force. Therefore, according to the work[53] compared to other PneuNet-based soft actuator without tendon, the tendon plays the role in stiffening the BISA.

In summary, the regulating the stiffness for BISA when the external load is against the bending direction is brought by the incremental air pressure in two forms: air pressure acting on contact areas and on lateral walls. During stiffening process, the contact area will also increase, enhancing its stiffness. For the external load that is applied in two different directions, against the bending and along the bending, The ATA can both help stiffen the actuator effectively.

## 3 Supplementary Note S3

The lateral stiffness is modulated through BLSs. When the bending is performed, the beam is subjected to bending and torsion. The curved beam is considered as the arc of a circle with a centre angle $\alpha$ and radius $R_{BLS}$ as the Figure 4B shows, which is the deformation characteristic of soft actuator when there is no load. The cruve of BLS can be expressed as:

$$\rho(\varphi) = R_{BLS}\varphi. \tag{3}$$

Following the two conditions that the material of BLS is linear-elastic and the stiffness is evaluated in the small displacement, the Castigliano's second theorem will be used to analyse the stiffness in this state. According to the Castigliano's second theorem, the displacement $\delta$ can be calculated by the strain energy satisfying the following equation:

$$\delta = \frac{\partial U_{BLS}}{\partial F_{ext}} \tag{4}$$

where $U_{BLS}$ is the strain energy of the BLS during deformation. Since the contribution of shear force to the strain energy is neglected, the strain energy can be determined as follows:

$$U_{BLS} = \frac{1}{EI}\int_0^\alpha M_{bending}^2 R_{BLS}\,\mathrm{d}\varphi + \frac{1}{GI_p}\int_0^\alpha M_{torsion}^2 R_{BLS}\,\mathrm{d}\varphi. \tag{5}$$

$M_{bending}$ and $M_{torsion}$ present the bending moment and torsional moment, which can be obtained as

$$\begin{cases} M_{bending} = F_{ext} R_{BLS} \sin(\alpha - \varphi) \\ M_{torsion} = F_{ext} R_{BLS} [1 - \cos(\alpha - \varphi)] \end{cases}. \tag{6}$$

In the bending process, the bending angle $\alpha$ and bending radius $R_{BLS}$ are not independent. It can be assumed that the arc length remains constant, formulated as

$$R_{BLS}\alpha = C \tag{7}$$

Based on equations (3) to (7), the PB-direction stiffness can be derived as

$$\frac{1}{k} = \frac{\delta}{F_{ext}} = C^3 \left[ \frac{1}{4EI} A_{bending}(\alpha) + \frac{1}{4GI_p} A_{torsion}(\alpha) \right] \tag{8}$$

where the $A_{bending}$ and $A_{torsion}$ are functions of the bending angle $\alpha_n$ and can represent the influence of bending and torsion. One obtains:

$$\begin{cases} A_{bending}(\alpha) = \dfrac{2}{\alpha^2} - \dfrac{\sin 2\alpha}{\alpha^3} \\ A_{torsion}(\alpha) = \dfrac{6}{\alpha^2} + \dfrac{\sin 2\alpha_n}{\alpha^3} - 8\dfrac{\sin\alpha}{\alpha^3} \end{cases}. \tag{9}$$

Except for the constants in the equation (8), it can be concluded that the stiffness is proportional to the value of the evaluation function $F(\alpha)$, which is

$$F(\alpha) = 1 \Big/ \left[ A_{bending}(\alpha) + \frac{2(1+\nu)A_{torsion}(\alpha)}{(1+\lambda^2)} \right] \tag{10}$$

where $\nu$ is the Poisson's ratio of the BLS's material and the $\lambda$ is the aspect ratio of the beam.

## 4 Supplementary Note S4

The grip consists of four fingers that surround the center connector, each finger is fixed to the center connector with a tilt of 15 degrees. In order to ensure the force is uniform in the process of grasping, the end of the finger is connected to a rope through a movable connector, which can adjust the bending degree of the finger. The each BLS of MTSA utilized the first control method, as Figure S2A shows. Each BLS was pretensioned to about 10N.

We divided the four fingers into two groups to control as Figure S6A shows, each group is separately driven by a steering gear (simplified diagram) pull rope and a pressure regulating valve to control air pressure. To make the grip structure compact, the steering gears of the two groups are arranged in the upper part and the lower part of the center connector respectively. As the Figure S6B-C illustrate, the Group1 rope bypasses the fixed pulley in the lower part and is connected to the steering gear fixed in the upper part. The Group2 rope is connected directly to the steering gear in the lower part.